\documentclass{article}

\usepackage[final]{neurips_2024}

\usepackage[utf8]{inputenc} 
\usepackage{enumitem}
\usepackage[T1]{fontenc}    
\usepackage{hyperref}       
\usepackage{url}            
\usepackage{booktabs}       
\usepackage{amsfonts}       
\usepackage{nicefrac}       
\usepackage{microtype}      
\usepackage{xcolor}         
\usepackage{listings}       
\usepackage{dirtytalk}
\usepackage{verbatim}
\usepackage{graphicx}       
\usepackage{wrapfig}

\title{Equitable Access to Justice: Logical LLMs Show Promise}

\vspace{-5mm}

\author{
    Manuj Kant\thanks{Equal contribution.} \\
    \texttt{kantmanuj@gmail.com}
    \And 
    Marzieh Nabi\footnotemark[1] \\
    \texttt{marzieh.nabi@gmail.com}
    \And
    Manav Kant\footnotemark[1] \\
    \texttt{mkant@caltech.edu}
    \And
    Preston Carlson \\
    \texttt{pjames27@stanford.edu}
    \And
    Megan Ma \\
    \texttt{meganma@law.stanford.edu}
}

\begin{document}

\vspace{-17mm}

\maketitle

\vspace{-7mm}

\begin{abstract}
\vspace{-2mm}
The costs and complexity of the American judicial system limit access to legal solutions for many Americans. Large language models (LLMs) hold great potential to improve access to justice. However, a major challenge in applying AI and LLMs in legal contexts, where consistency and reliability are crucial, is the need for System 2 reasoning. In this paper, we explore the integration of LLMs with logic programming to enhance their ability to reason, bringing their strategic capabilities closer to that of a skilled lawyer. Our objective is to translate laws and contracts into logic programs that can be applied to specific legal cases, with a focus on insurance contracts. We demonstrate that while GPT-4o fails to encode a simple health insurance contract into logical code, the recently released OpenAI o1-preview model succeeds, exemplifying how LLMs with advanced System 2 reasoning capabilities can expand access to justice.

\end{abstract}

\vspace{-6mm}

\section{Introduction}
\vspace{-2mm}
Access to legal solutions has become increasingly limited across the low, middle, and upper-middle classes, with all facing significant barriers. More than $75\%$ of litigants represent themselves \cite{Court_stats}, with California alone reporting over $4.3$ million self-represented litigants \cite{Self_rep_CA}. This trend is largely attributed to the high cost of legal services and widespread distrust of attorneys, as indicated by surveys conducted by the California judicial system \cite{CA_Council}. Addressing these challenges requires the development of reliable and transparent technological solutions to bridge the considerable gaps in the legal system for consumers.


Recently, legal applications have garnered significant attention as a promising use case for LLMs. Several scientific studies and business initiatives have highlighted both the potential and limitations of LLMs in the legal domain \cite{hallucinations_legal}.
Considerable progress is still required before these technologies can deliver consistent and transparent solutions. While human lawyers can articulate the reasoning behind their decisions and strategies, LLMs currently lack this capability to a sufficient degree. \cite{chainofthoughs_2023, Faithful_CoT, GoT_2024, Reasoning_survey, Bootstrapping_reasoning, ToT_2023}.

Before discussing which AI solutions might be appropriate for legal applications, it is crucial to first consider whether law is inherently deterministic. While laws provide a structured framework that can seem deterministic, the human element, interpretation, judgment, and discretion introduce a degree of uncertainty, making the law not entirely deterministic in practice. In summary, the deterministic aspects of law includes a) legal rules and statutes; and b) case precedents. The non-deterministic aspects are a) judicial interpretation; b) human judgment; and c) equity and fairness. 



Given the multifaceted nature of legal practices, we propose that a combination of probabilistic and deterministic AI solutions is required to effectively address legal planning and reasoning. This raises the next logical question: which AI algorithms and relational frameworks are most suitable for developing reliable legal assistance? 
In the following sections of this article, we outline our current approach to addressing these challenges, supported by an experiment that illustrate our findings in the context of contracts. We will then discuss our broader strategy and the future directions we intend to explore.

\section{Our Current Approach}
\vspace{-2mm}
LLMs are great probabilistic solutions with rapid improvements in their capabilities. However, given their inherent probabilistic nature, there is always a chance of hallucination and inconsistent answers. On the other hand, we have logic programs with highly consistent responses and explainable answers. But their main limitation lies in their inherent lack of flexibility and scalability for handling certain complex tasks, especially in real-world applications as they struggle to model uncertainty, probabilistic reasoning, or temporal dynamics. 

To leverage the strengths of both LLMs and logic programming, we explore various hybrid approaches that combine these two methodologies. In one such approach, LLMs are employed to automatically generate logical representations of legal statutes or rules. Once these representations are constructed, the specific details of a given case can be applied to this logic-based framework. This allows for a structured reasoning process, where the law's application to individual cases is derived through formal logic, thereby enhancing the interpretability and precision of legal decision-making.

The integration of LLMs with logic programming (neuro-symbolic AI) is becoming increasingly popular. AlphaGeometry \cite{alpha_geometry} is a great example of the new horizons achievable by leveraging the strengths of each method. In \cite{Berkley_2024_Neurosym}, the authors proposed a neuro-symbolic approach, leveraging LLMs to generate logical representations of problems, with Prolog handling the deductive reasoning. 


\vspace{-2mm}
\subsection{ Limitations of Our Current Approach}
\vspace{-2mm}
In our current approach, we leverage LLMs to generate logical representations from legal texts. LLMs offer a significant advantage in developing these representations at scale, enabling the efficient processing of vast and complex legal corpora. However, the accuracy and quality of the logic produced by LLMs remain a critical concern, as these models can misinterpret legal terms, omit critical details, generate logical inconsistencies, or overgeneralize legal principles. Additionally, LLMs may struggle with nuances, ambiguities, and the conditional or temporal relationships inherent in legal texts, leading to potential errors. Moreover, potential biases in their original training data can further compromise the validity of the generated logic. Therefore, it is essential to implement robust mechanisms to prevent these types of errors and mitigate the potential negative impact of LLMs. Ensuring the integrity of the generated logic is crucial for maintaining the reliability and trustworthiness of our proposed approach.

Encouragingly, however, we have found that the quality of the logical representations generated by LLMs is significantly improving as these models become more powerful and sophisticated. In the Experiment section, we demonstrate the quality differences between two of the most recent OpenAI models, GPT-4o and OpenAI o1-preview, specifically in generating Prolog representations of certain legal contracts. Another important mechanism to ensure the accuracy of these logical representations is incorporating human feedback. To achieve this, we propose having expert attorneys, familiar with the specific legal domains, review the generated logic to validate and further enhance its quality.
\vspace{-2.0mm}

\section{Experiment - Hospital Cash Benefit Policy}
\vspace{-2.0mm}



In our experiments, we focus on legal contracts, particularly the challenges consumers face in understanding health insurance coverage. A June 2024 Stanford survey revealed that $83\%$ of participants used traditional methods to check their insurance policy, with $82\%$ finding the process frustrating. Computational law experts highlight the importance of "computable contracts" \cite{computable_commercial_contracts} to reduce confusion and help identify coverage gaps.

Computable Contracts: Ideally, insurance contracts would be represented as interpretable computer programs (computable contracts), which could be easily audited by regulatory bodies and legal/domain experts. These programs would allow consumers to check their coverage through a simple command rather than poring over complex documents. Logic programming languages like Prolog, which exhibit logical reasoning, offer both interpretability and automation.

However, manually encoding contracts into logic programs is time-consuming and not scalable. We demonstrate how LLMs can assist in scaling this encoding process. Specifically, the recent OpenAI o1-preview model, with advanced System 2 reasoning capabilities, significantly outperforms its predecessor, GPT-4o, in encoding policies into logic programs. In essence, we show that advanced System 2 reasoning in LLMs enables the creation of logic programs with similarly advanced reasoning abilities.

\textbf{Approach} - We prompted both GPT-4o and OpenAI o1-preview (see Appendix~\ref{app:policy_encoding_prompt} for the prompt) to translate a simplified version of the Chubb Hospital Cash Benefit insurance policy \cite{chubb_simplified} (Appendix~\ref{app:simplified_chubb}) into logical rules in Prolog. The prompt provided no hints or guidelines regarding code structure, but provided a couple of baseline assumptions and clarified that the code should answer questions about whether a given claim is covered under the policy. Note that all prompts to {\it both} GPT-4o and OpenAI o1-preview in this paper were made with the default parameters for o1-preview (top\_p = 1, temperature = 1, n = 1, presence\_penalty = 0, frequency\_penalty = 0).

\begin{wrapfigure}{l}{0.5\textwidth}   
\centering
\vspace{-3mm}
\includegraphics[width=.7\linewidth, height=7.2cm]{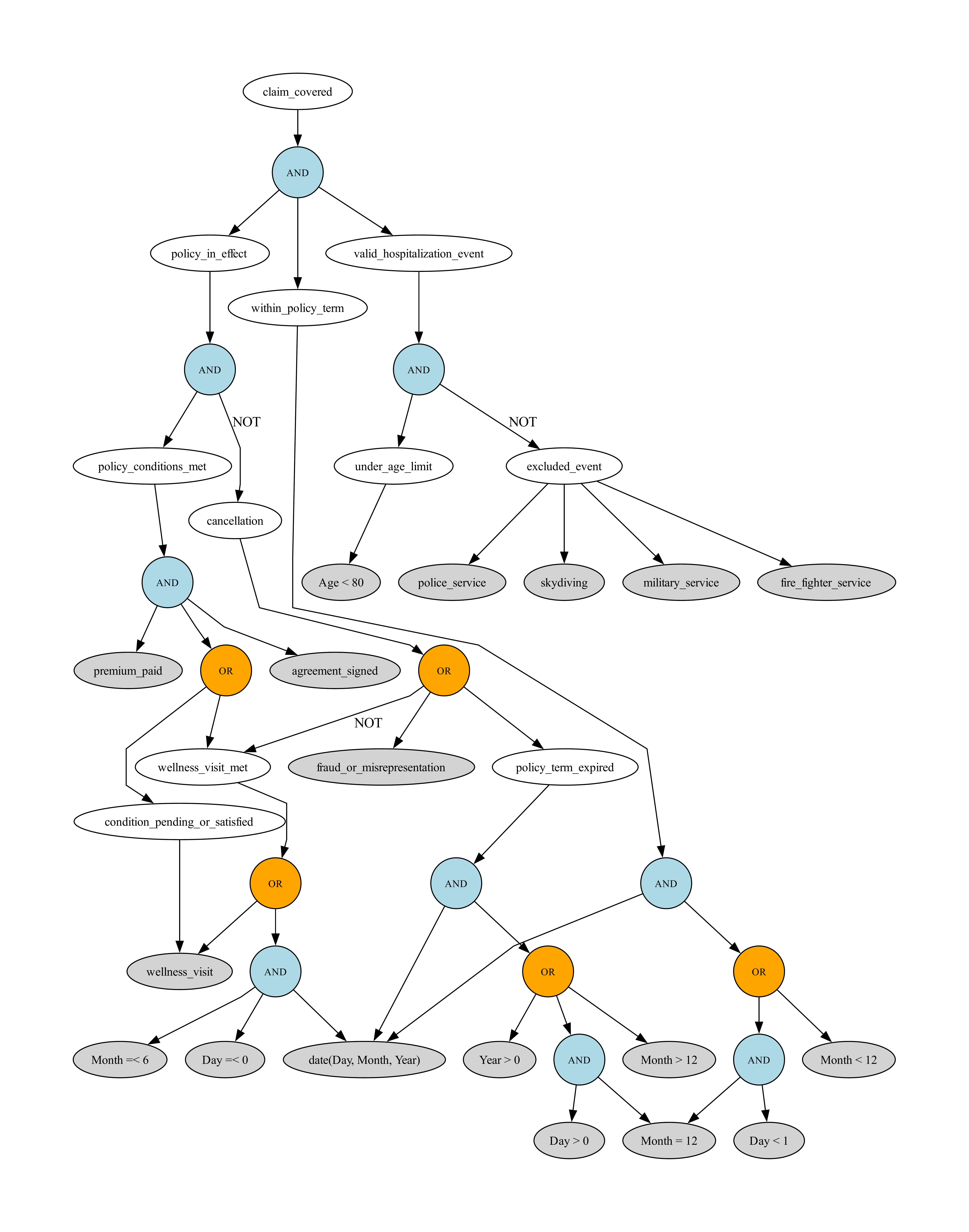}
\vspace{-2mm}
\small 
\caption{\label{fig:4o-simplified-chubb} \small Prolog translation of simplified Chubb policy generated by GPT-4o. Corresponding code in Appendix~\ref{app:4o_policy_encoding}}
\vspace{-3mm}
\end{wrapfigure}

\begin{wrapfigure}{r}{0.6\textwidth}   
\centering
\vspace{-5mm}
\includegraphics[width=0.65\textwidth, height=5cm]{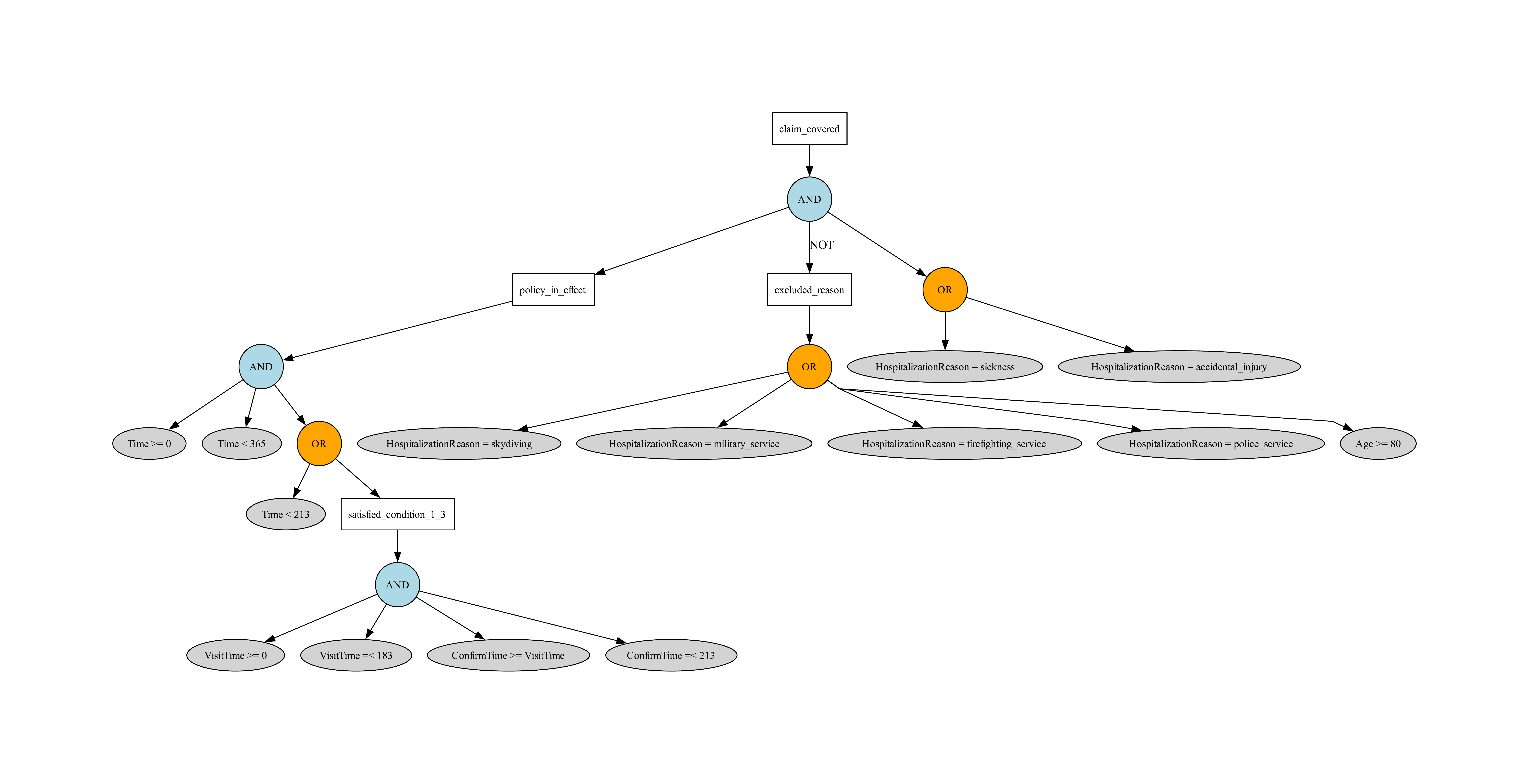}
\vspace{-10mm}
\caption{\label{fig:o1-simplified-chubb} \small Prolog translation of simplified Chubb policy generated by OpenAI o1-preview. Corresponding code in Appendix~\ref{app:o1_policy_encoding}}
\vspace{-3mm}
\end{wrapfigure}

Graphical representations of the code output generated by each model are shown in Figure~\ref{fig:4o-simplified-chubb} and Figure~\ref{fig:o1-simplified-chubb}. Both models structured the code with a single root node, claim\_covered, determining whether a claim is covered.

The comparison between the two encodings focuses on the condition in Section $1.3$, which requires that {\it no later than the 7th month anniversary of the policy's effective date, written confirmation from the medical provider regarding a wellness visit within 6 months must be supplied.} Additionally, for the policy to remain in effect, {\it the condition in Section 1.3 must either be pending or satisfied in a timely manner.}

This means that, within 7 months of the policy start date, the insuree must provide confirmation of a wellness visit made within the first 6 months. However, if 7 months have not yet passed, the condition is considered \say{pending,} and the policy can remain active without confirmation of the wellness visit.

\textbf{GPT-4o Policy Encoding Analysis} - A cursory look at the leaves of GPT-4o's encoding reveals its lack of understanding of this condition, as there is simply no node checking for whether some action is taken within $7$ months (although there are conditions checking for each of $6$ and $12$ months).

Taking a deeper look at the mechanisms by which GPT-4o attempts to encode the wellness visit condition, we see (tracing the graph) that one of condition\_pending\_or\_satisfied and wellness\_visit\_met are required for policy\_in\_effect. Then, condition\_pending\_or\_satisfied is satisfied exactly when wellness\_visit is. This seems to indicate that a wellness visit is required for the condition to be satisfied OR pending, which is odd since if the condition is still pending, then the wellness visit should not be required. Furthermore, wellness\_visit\_met can also be satisfied by wellness\_visit OR if date (which presumably refers to the date of the wellness visit) is at most $6$ months.

One should feel no shame in finding the above confusing, as the delineated logic is not only incorrect in several ways, but also muddled and disorganized. The \say{pending} part of the condition is omitted from the encoding in all but name, the wellness\_visit node is redundantly used by both condition\_pending\_or\_satisfied and wellness\_visit\_met, and the semantic meanings of some variables and nodes are murky at best. Even on this simplified version of a relatively simple contract, GPT-4o's encoding is a far cry from the interpretable and automated ideal of computable contracts.

\textbf{OpenAI o1-preview Policy Encoding Analysis} - Even at first glance, the policy encoding generated by OpenAI o1-preview looks much more organized than the one generated by GPT-4o.

Focusing again on how the model encodes the condition defined in Section $1.3$, we first note that there are three relevant temporal variables in this encoding: Time, VisitTime, and ConfirmTime. These correspond with the time of hospitalization (i.e. when the claim is made), the time of the required wellness visit, and the time at which confirmation of that visit was given to the insurer, respectively. Then, tracing the graph, we see that Time $< 213$ or satisfied\_condition\_1\_3 for policy\_in\_effect, where this encoding seems to be measuring time in days. 

 Thus, either the time of hospitalization is within the $7$ month period in which the condition defined in Section $1.3$ is considered pending, or that condition must be satisfied. For the condition to be satisfied, it is required that $0 \leq $ VisitTime $ \leq 183$ and VisitTime $ \leq $ ConfirmTime $ \leq 213$. Thus, the wellness visit must have occurred within $6$ months and the confirmation given within $7$.

The simple and organized manner in which OpenAI o1-preview encodes the above condition is a testament to its advanced reasoning capabilities. 

\textbf{Empirical Comparison of Policy Encoding Correctness} - While our graphical representations indicate that o1-preview’s encoding of the contract is more interpretable than 4o’s, we also conducted an empirical measurement to compare the models' respective accuracies. We performed ten trials comparing o1-preview's and 4o’s encodings of the simplified Chubb insurance policy. In each trial, both models generated their own policy encoding (Appendix~\ref{app:llm_policy_encodings}) from the prompt in Appendix~\ref{app:policy_encoding_prompt}. Then, both models were prompted (Appendix~\ref{app:query_encoding_prompt}) to translate nine natural language yes/no questions (Appendix~\ref{app:queries_and_answers}) into Prolog queries (Appendix~\ref{app:llm_query_encodings}) on their respective policy encodings. Using SWISH \cite{swish}, the policy code from each model was run on the query encodings generated for that policy code, and the number of correct answers was recorded.

\vspace{-1mm}
\begin{small}
\begin{table}[ht]
\centering
\resizebox{1\linewidth}{!}{%
\begin{tabular}{|c|c|c|c|c|c|c|c|c|c|c|c|}
\hline
 & Trial 1 & Trial 2 & Trial 3 & Trial 4 & Trial 5 & Trial 6 & Trial 7 & Trial 8 & Trial 9 & Trial 10 & Mean \\ \hline
o1-preview & 8 & 7 & 8 & 8 & 8 & 8 & 6 & 8 & 6 & 8 & 7.5 ± 0.23 \\ \hline
4o & 6 & 0 & 5 & 2 & 0 & 0 & 5 & 5 & 0 & 1 & 2.4 ± 0.81 \\ \hline
\end{tabular}%
}
\caption{\small Scores of o1-preview and GPT-4o across trials and their means (Error bars calculated via calculator represent 1-sigma sample standard error over trial-to-trial variability, assuming normal errors.).}
\end{table}
\end{small}
\vspace{-5mm}
On average, 4o answered $2.4$ out of $9$ queries correctly, while o1-preview averaged $7.5$. In four of the trials, 4o scored $0$ (due to bad syntax in the policy encoding). Even when excluding these runs, 4o averaged only $4$ correct answers, $3.5$ fewer than o1-preview. This demonstrates that o1-preview’s queries on its policy encodings yielded much more accurate results than those of 4o.

\vspace{-2mm}
\section{Our Future Approaches}
\vspace{-2mm}
We are on the cusp of an exciting era where AI can make legal solutions more accessible by applying human-like thinking, including planning and reasoning. In addition to our approach in this paper--using LLMs to generate logical representations--we explore several other potential approaches.

One approach within the realm of LLMs and logic programming is to fine-tune language models using logic-based explanations. In \cite{Bootstrapping_reasoning}, the authors demonstrated how the "Self-Taught Reasoner (STaR)" method enhances language model reasoning through rationale generation, which provides step-by-step explanations, and rationalization, which corrects incorrect answers. This iterative process improves reasoning capabilities without the need for large annotated datasets. Applying a similar approach with logic generation could enhance the legal reasoning capabilities of language models

Our second proposal focuses on generating knowledge graph representations of legal systems, similar to the mental models experienced attorneys develop. These models integrate local and federal laws, case precedents, and relevant facts, as well as more nuanced factors. We propose using LLMs to create local knowledge graphs that mirror these mental models. Each jurisdiction would have its own tailored graph, which experts would review and refine, enabling more accurate legal decision-making.

Our third proposal suggests a dynamic programming approach to legal planning using LLMs and logic-based representations of the law. Like experienced attorneys who continuously refine legal strategies based on new information, this approach involves simulating scenarios, assessing risks, and adjusting plans iteratively. We propose digitizing this process with LLMs, generating logic-based representations of laws, statutes, and precedents \cite{dynamic_planning_2023}. Each decision will build on prior steps, allowing for continuous refinement and optimization of legal strategies in a digital environment.

\newpage
\newpage
\appendix
\section{Appendix / supplemental material}

\subsection{Simplified Chubb Hospital Cash Benefit Policy}
\label{app:simplified_chubb}
\noindent Between:\\
CODEX INSURANCE LIMITED (\textquotedblleft us\textquotedblright)\\
and\\
\_\_\_\_\_\_\_\_\_\_\_\_\_\_\_\_ (\textquotedblleft You\textquotedblright)\\

\vspace{1em}
This policy is provided on the following terms and conditions:

POLICY IN EFFECT AND CONDITIONS

1.1 The payment of any benefit under this policy is conditioned on the policy being in effect at the time of the hospitalization for sickness or accidental injury on which the claim for such benefit is premised. The policy will be in effect if:
\begin{enumerate}[label=(\alph*)]
    \item This agreement is signed,
    \item The applicable premium for the policy period has been paid, and
    \item The condition set out in Section 1.3 is still pending or has been satisfied in a timely fashion, and
    \item The policy has not been canceled.
\end{enumerate}

1.2 Cancelation will be deemed to have occurred if there is fraud, or any misrepresentation or material withholding of any information provided by you to the Company in connection with any communication or information relating to this policy, or if the condition set out in Section 1.3 has not been satisfied in a timely fashion. It will also be automatically canceled at midnight, US Eastern time then in effect, on the last day of the policy term described in Section 5 below.

1.3 No later than the 7th month anniversary of the effective date of this policy, you will supply us with written confirmation from the medical provider in question of a wellness visit for yourself with a qualified medical provider occurring no later than the 6th month anniversary of the effective date of this policy.

GENERAL EXCLUSIONS

2.1 Your policy will not apply to, and no benefit will be paid with respect to, any event causing sickness or accidental injury arising directly or indirectly out of:
\begin{enumerate}
    \item Skydiving; or
    \item Service in the military; or
    \item Service as a fire fighter; or
    \item Service in the police; or
    \item If your age at the time of the hospitalization is equal to or greater than 80 years of age.
\end{enumerate}

GENERAL CONDITIONS

3.1 Where does Your Policy apply?

3.1.1 Your Policy insures You twenty-four (24) hours a day anywhere in the world.

3.2 Arbitration

3.2.1 If any dispute or disagreement arises regarding any matter pertaining to or concerning this Policy, the dispute or disagreement must be referred to arbitration in accordance with the provisions of the Arbitration Act (Cap. 10) and any statutory modification or re-enactment thereof then in force, such arbitration to be commenced within three (3) months from the day such parties are unable to settle the dispute or difference. If You fail to commence arbitration in accordance with this clause, it is agreed that any cause of action and any right to make a claim that You have or may have against Us shall be extinguished completely. Where there is a dispute or disagreement, the issuance of a valid arbitration award shall also be a condition precedent to our liability under this Policy. In no case shall You seek to recover on this Policy before the expiration of sixty (60) days after written proof of claim has been submitted to Us in accordance with the provisions of this Policy.

3.3 Laws of New York

3.3.1 Your Policy is governed by the laws of New York.

3.4 US Currency

3.4.1 All payments by You to Us and by Us to You or someone else under your policy must be in United States currency.

3.5 Premium

3.5.1 The premium described in Section 5 below shall be paid in one lump sum at the signing of the policy.

3.6 Policy Term
The term of this policy will begin on the date accepted by Us as signified by our signature of the policy (the effective date) and will last for a period of one year from that date, unless previously canceled pursuant to Section 1 above.

\subsection{Prompt for Generating Policy Encodings}
\label{app:policy_encoding_prompt}
- Given the insurance contract below, translate the document into valid Prolog rules so that I can run a Prolog query on the code regarding whether or not some claim is covered under the policy and receive the correct answer to the question.

- Please fully define all predicates and DO NOT define any facts, only rules that can be used to answer queries on this insurance contract.

- Assume that all dates/times in any query to this code (apart from the claimant's age) will be given RELATIVE to the effective date of the policy (i.e. there will never be a need to calculate the time elapsed between two dates). Take dates RELATIVE TO the effective date into account when writing this encoding.

- Assume that the agreement has been signed and the premium has been paid (on time). There is no need to encode rules or facts for these conditions.

- Return only Prolog code in your reply. No explanation is necessary.

- Ensure that:
\begin{enumerate}
\item The legal text is appropriately translated into correct Prolog rules.
\item The output does not redefine, misuse, or conflict with any built-in Prolog predicates.
\item If dynamic predicates are necessary, they are declared and managed correctly.
\item All predicates used in the generated Prolog code, including those referenced in the query, are fully defined and error-free to prevent issues like "procedure does not exist."
\item Logical relationships, conditions, and dependencies in the text are faithfully represented in the Prolog rules to ensure accurate query results.
\end{enumerate}

- Insurance contract:
\{text\_content\}

\subsection{LLM Policy Encodings}
\label{app:llm_policy_encodings}
\subsubsection{GPT-4o Policy Encoding}
\label{app:4o_policy_encoding}
\begin{lstlisting}
% Dynamic predicates to allow for updates if necessary
:- dynamic policy_in_effect/0.
:- dynamic cancellation/0.
:- dynamic wellness_visit/0.

% Policy will be in effect if it has not been canceled and all conditions are met.
policy_in_effect :-
    \+ cancellation,
    policy_conditions_met.

% Conditions for the policy to be in effect
policy_conditions_met :-
    agreement_signed,
    premium_paid,
    (condition_pending_or_satisfied; wellness_visit_met).

% Policy cancellation conditions
cancellation :-
    fraud_or_misrepresentation,
    !.
cancellation :-
    \+ wellness_visit_met,
    !.
cancellation :-
    policy_term_expired.

% Condition 1.3
condition_pending_or_satisfied :- wellness_visit.

% Policy term expiration check
policy_term_expired :-
    date(Day, Month, Year),
    (Month > 12; (Month = 12, Day > 0) ; Year > 0).

% Placeholder predicates to simulate conditions
agreement_signed.
premium_paid.
fraud_or_misrepresentation :- fail. % No fraud by default

% Wellness visit check
wellness_visit_met :-
    date(Day, Month, _),
    Month =< 6,
    Day =< 0 ; % Assuming verification by certain day or prior month with Day <= 0
    wellness_visit.

% Check if a claim is covered
claim_covered(Accident, Hospitalization, Age) :-
    policy_in_effect,
    within_policy_term,
    valid_hospitalization_event(Accident, Age).

% Validity check for hospitalization event
valid_hospitalization_event(Accident, Age) :-
    \+ excluded_event(Accident),
    under_age_limit(Age).

% Exclusions based on events
excluded_event(skydiving).
excluded_event(military_service).
excluded_event(fire_fighter_service).
excluded_event(police_service).

% Age check
under_age_limit(Age) :-
    Age < 80.

% Check if within policy term
within_policy_term :-
    date(Day, Month, _),
    (Month < 12 ; (Month = 12, Day < 1)).

% Placeholder date predicate
date(0, 0, 0). % Assuming effective date being the 0th day of the 0th month

% Queries can be structured as:
% ?- claim_covered(Accident, Hospitalization, Age).
% where Accident is one of [skydiving, military_service, fire_fighter_service, police_service, other],
% and Age is a valid integer representing the claimant's age.
\end{lstlisting}

\subsubsection{OpenAI o1-preview Policy Encoding}
\label{app:o1_policy_encoding}

\begin{lstlisting}

policy_in_effect(Time, VisitTime, ConfirmTime) :-
    Time >= 0,
    Time < 365,
    (Time < 213 ; satisfied_condition_1_3(VisitTime, ConfirmTime)).

satisfied_condition_1_3(VisitTime, ConfirmTime) :-
    VisitTime >= 0,
    VisitTime =< 183,
    ConfirmTime >= VisitTime,
    ConfirmTime =< 213.

claim_covered(Time, VisitTime, ConfirmTime, HospitalizationReason, Age) :-
    policy_in_effect(Time, VisitTime, ConfirmTime),
    (HospitalizationReason = sickness ; HospitalizationReason = accidental_injury),
    \+ excluded_reason(HospitalizationReason, Age).

excluded_reason(HospitalizationReason, Age) :-
    HospitalizationReason = skydiving ;
    HospitalizationReason = military_service ;
    HospitalizationReason = firefighting_service ;
    HospitalizationReason = police_service ;
    Age >= 80.
\end{lstlisting}



\subsection{Queries and Answers for Empirical Evaluation}
\label{app:queries_and_answers}
All queries are preceded by the disclaimer: \say{Assuming all other conditions are met and no other exclusions apply (where by 'other,' I mean anything not referenced in the query that follows),\ldots}

\textbf{Query: } \say{will the claim be covered under the policy if I was hospitalized by burns suffered while doing my duty as a firefighter?}
\textbf{Answer: } \say{No.}

\textbf{Query: } \say{will the claim be covered under the policy if I am 78 years old at the time of hospitalization?}
\textbf{Answer: } \say{Yes.}

\textbf{Query: } \say{will the claim be covered under the policy if I was hospitalized for pneumonia 5 months after the policy's effective date, and my age at the time of hospitalization is 65?}
\textbf{Answer: } \say{Yes.}

\textbf{Query: } \say{will the claim be covered under the policy if I was hospitalized due to a fall while traveling abroad and I had given confirmation of my wellness visit 8 months after the policy's effective date?}
\textbf{Answer: } \say{No.}

\textbf{Query: } \say{will the claim be covered under the policy if I was hospitalized for punching my own face to show off for my friends and I did not commit fraud or misrepresentation?}
\textbf{Answer: } \say{No.}

\textbf{Query: } \say{will the claim be covered under the policy if I was hospitalized due to an injury sustained while skydiving, my age at the time of hospitalization was 79, and proof of my wellness visit was provided 6.5 months after the policy's effective date?}
\textbf{Answer: } \say{No.}

\textbf{Query: } \say{will the claim be covered under the policy if I was hospitalized for a heart attack, proof of the wellness visit was submitted 2 months after the policy's effective date, and my age at the time of hospitalization was 75?}
\textbf{Answer: } \say{Yes.}

\textbf{Query: } \say{will the claim be covered under the policy if I was hospitalized after being injured in a military training exercise, the hospitalization occurred within the policy term, and I did not commit fraud?}
\textbf{Answer: } \say{No.}

\textbf{Query: } \say{will the claim be covered under the policy if I was hospitalized due to my son biting me in the ankle, proof of my wellness visit was provided 6 months after the effective date, and I was serving as a police officer at the time of hospitalization?}
\textbf{Answer: } \say{Yes.}

\subsection{Prompt for Generating Query Encodings}
\label{app:query_encoding_prompt}
- I have given below:
\begin{enumerate}
\item A question about whether or not the policy defined in a given insurance contract applies in a particular situation
\item The text of the insurance contract
\item A Prolog encoding of the insurance contract
\end{enumerate}
- Encode the question into a Prolog query such that it can be run on the Prolog encoding of the insurance contract, returning the correct answer to the question.

- Assume that the agreement has been signed and the premium has been paid (on time). There is no need to encode rules or facts for these conditions.

- Return only Prolog code in your reply. No explanation is necessary.

- Ensure that:
\begin{enumerate}
\item The output does not redefine, misuse, or conflict with any built-in Prolog predicates.
\item If dynamic predicates are necessary, they are declared and managed correctly.
\item All predicates used in the generated Prolog code, including those referenced in the query, are fully defined and error-free to prevent issues like "procedure does not exist."
\item Logical relationships, conditions, and dependencies in the text are faithfully represented in the Prolog rules to ensure accurate query results.
\item No absolute dates/times (apart from the claimant's age) are encoded in your query. Only include dates/times RELATIVE to the effective date of the policy (again, except for age).
\item Set any facts/rules/parameters in the code such that ALL conditions (for the policy to apply) which are UNRELATED to the above query are satisfied.
\item Set any facts/rules/parameters in the code such that NO exclusions (which would prevent the policy from applying) which are UNRELATED to the above query are satisfied.
\end{enumerate}

- Question:
\{query\}

- Insurance contract: 
\{text\_content\}

- Insurance contract Prolog encoding:
\{policy\_encoding\}

\subsection{LLM Query Encodings}
\label{app:llm_query_encodings}
\subsubsection{GPT-4o Query Encodings}
\label{app:4o_query_encodings}
\begin{lstlisting}
Query 1 Prolog:
?- claim_covered(fire_fighter_service, hospitalization, Age).
\end{lstlisting}

\begin{lstlisting}
Query 2 Prolog:
?- claim_covered(other, hospitalization, 78).
\end{lstlisting}

\begin{lstlisting}
Query 3 Prolog:
?- claim_covered(other, true, 65).
\end{lstlisting}

\begin{lstlisting}
Query 4 Prolog:
% Updating the current conditions for the query
:- retractall(date(_, _, _)).
:- assert(date(0, 8, 0)). % 8 months after policy effective date

% Mark health visit as complete, despite late confirmation
:- retractall(wellness_visit).
:- assert(wellness_visit).

% Formatting the query based on the claim details
?- claim_covered(other, hospitalization, 40). % Fall during travel, hospitalized, age is 40
\end{lstlisting}

\begin{lstlisting}
Query 5 Prolog:
% Define the incident with its characteristics
incident_details(punching_own_face, hospitalized, 25). % Example age, assuming it's 25

% Prolog query to determine if the claim is covered
?- incident_details(Accident, Hospitalization, Age),
   claim_covered(Accident, Hospitalization, Age).
\end{lstlisting}

\begin{lstlisting}
Query 6 Prolog:
% Set up the situation to reflect the claim question
date(7, 0, 0). % Assume the date for checking purposes

% Query to check if the claim is covered
?- claim_covered(skydiving, true, 79).
\end{lstlisting}

\begin{lstlisting}
Query 7 Prolog:
% Prolog query for the given scenario
?- assert(date(0, 2, 0)), % Setting date to 2 months after policy effective date
   claim_covered(other, heart_attack, 75).
\end{lstlisting}

\begin{lstlisting}
Query 8 Prolog:
?- claim_covered(military_service, Hospitalization, Age).
\end{lstlisting}

\begin{lstlisting}
Query 9 Prolog:
?- claim_covered(other, hospitalization, 30).
\end{lstlisting}

\subsubsection{OpenAI o1-preview Query Encodings}
\label{app:o1_query_encodings}
\begin{lstlisting}
Query 1 Prolog:
policy_in_effect(Time, VisitTime, ConfirmTime) :-
    Time >= 0,
    Time < 365,
    (Time < 213 ; satisfied_condition_1_3(VisitTime, ConfirmTime)).

satisfied_condition_1_3(VisitTime, ConfirmTime) :-
    VisitTime >= 0,
    VisitTime =< 183,
    ConfirmTime >= VisitTime,
    ConfirmTime =< 213.

excluded_reason(Cause, Age) :-
    Cause = skydiving ;
    Cause = military_service ;
    Cause = firefighting_service ;
    Cause = police_service ;
    Age >= 80.

claim_covered(Time, VisitTime, ConfirmTime, HospitalizationReason, Cause, Age) :-
    policy_in_effect(Time, VisitTime, ConfirmTime),
    (HospitalizationReason = sickness ; HospitalizationReason = accidental_injury),
    \+ excluded_reason(Cause, Age).

?- claim_covered(100, 150, 160, accidental_injury, firefighting_service, 40).
\end{lstlisting}

\begin{lstlisting}
Query 2 Prolog:
claim_covered(200, 150, 160, sickness, 78).
\end{lstlisting}

\begin{lstlisting}
Query 3 Prolog:
?- claim_covered(150, 0, 0, sickness, 65).
\end{lstlisting}

\begin{lstlisting}
Query 4 Prolog:
claim_covered(240, 180, 244, accidental_injury, 30).
\end{lstlisting}

\begin{lstlisting}
Query 5 Prolog:
?- claim_covered(100, 0, 0, intentional_injury, 30).
\end{lstlisting}

\begin{lstlisting}
Query 6 Prolog:
?- claim_covered(200, 180, 198, skydiving, 79).
\end{lstlisting}

\begin{lstlisting}
Query 7 Prolog:
?- claim_covered(100, 30, 60, sickness, 75).
\end{lstlisting}

\begin{lstlisting}
Query 8 Prolog:
?- claim_covered(100, 0, 0, military_service, 40).
\end{lstlisting}

\begin{lstlisting}
Query 9 Prolog:
?- claim_covered(200, 180, 183, accidental_injury, 40).
\end{lstlisting}

\newpage

\end{document}